\documentclass{bmvc2k}
\usepackage{graphicx}
\usepackage{epstopdf}
\usepackage{booktabs}
\usepackage{makecell}
\usepackage{amssymb}
\usepackage{pifont}
\usepackage{multirow}
\usepackage{xcolor}
\usepackage{comment}
\usepackage{threeparttable}

\title{Learning Separable Hidden Unit Contributions for Speaker-Adaptive Lip-Reading}

\addauthor{Songtao Luo}{luosongtao18@mails.ucas.ac.cn}{1,2}
\addauthor{Shuang Yang}{shuang.yang@ict.ac.cn}{1,2}
\addauthor{Shiguang Shan }{sgshan@ict.ac.cn}{1,2}
\addauthor{Xilin Chen}{xlchen@ict.ac.cn}{1,2}
\addinstitution{
Key Laboratory of Intelligent\\
Information Processing of Chinese\\
Academy of Sciences (CAS),\\
Institute of Computing Technology,\\
CAS,\\
Beijing, 100190, China
}
\addinstitution{
University of Chinese Academy of\\
Sciences,\\
Beijing, 100049, China
}

\runninghead{LUO ET AL.}{Learning Separable Hidden Unit Contributions}

\begin{document}

\maketitle
\vspace{-2em}
\begin{abstract}
In this paper, we propose a novel method for speaker adaptation in lip reading, motivated by two observations.
Firstly, a speaker's own characteristics can always be portrayed well by his/her few facial images or even a single image with shallow networks, while the fine-grained dynamic features associated with speech content expressed by the talking face always need deep sequential networks to represent accurately. Therefore, we treat the shallow and deep layers differently for speaker adaptive lip reading.  
Secondly, we observe that a speaker's unique characteristics ( e.g. prominent oral cavity and mandible) have varied effects on lip reading performance for different words and pronunciations, necessitating adaptive enhancement or suppression of the features for robust lip reading. 
Based on these two observations, we propose to take advantage of the speaker's own characteristics to automatically learn separable hidden unit contributions with different targets for shallow layers and deep layers respectively. 
For shallow layers where features related to the speaker's characteristics are stronger than the speech content related features, we introduce speaker-adaptive features to learn for enhancing the speech content features. 
For deep layers where both the speaker's features and the speech content features are all expressed well, we introduce the speaker-adaptive features to learn for suppressing the speech content irrelevant noise for robust lip reading. 
Our approach consistently outperforms existing methods, as confirmed by comprehensive analysis and comparison across different settings.
Besides the evaluation on the popular LRW-ID and GRID datasets, we also release a new dataset for evaluation, CAS-VSR-S68, to further assess the performance in an extreme setting where just a few speakers are available but the speech content covers a large and diversified range. The results demonstrated our method's superiority on this challenging dataset as well.
\end{abstract}

\vspace{-1em}
\section{Introduction}
\vspace{-1em}
\label{Introduction}
\par
Lip reading, or Visual Speech Recognition (VSR), is a challenging task that aims to interpret the spoken content by analyzing visual cues of a speaker's lip or face movements. 
%
Thanks to the emergence of large-scale lip reading datasets \cite{yang2019lrw,Chung17,afouras2018deep,afouras2018lrs3,Chung16} and pre-training methods \cite{shi2022robust,lian2023av,haliassos2022jointly,baevski2022data2vec}, the domain of lip reading has achieved significant progress recently. Currently, the performance of lip reading has reached a level where it can rival the performance of audio-based speech recognition models from four years ago\cite{chang2023conformers}. 
However, lip reading still encounters various challenges, especially the diverse speaking styles and facial appearance of different speakers, which severely limits the practical application of lip reading in the real world.


By observing a speaker's characteristics, we find that most of the speaker's characteristics, such as his/her facial structure, mouth shape, skin texture, skin tone, beard, glasses, markings, and other facial traits, are usually static and would not be significantly affected by pronouncing different words. This property enables the corresponding speaker to be easily be  distinguished by checking the speaker's few facial images or even a single image with shallow networks.
However, when we are speaking, our facial region, especially the lip region, is always in a state of motion during the whole speaking process. The dynamics involved in the process of speaking are typically fine-grained spatio-temporal changes, which require deep sequential networks to obtain good representations.

On the other hand, although most of the speaker's characteristics, are static and not affected by pronouncing different speech words, there exist some attributes which can enhance or weaken the facial dynamic information during speech production. 
For example, for a speaker with lips that turn outward, compared to others, the lip movement for plosive sounds will be weakened (due to the relatively small lip movement distance), while the lip movement for fricative sounds will be enhanced (due to relatively large lip movement distance).
Therefore, we propose to leverage the speaker's characteristics for learning to enhance and suppress separate hidden unit contributions in the shallow layers and deep layers respectively.


\begin{figure*}
    \subfigure[Overall Growth Trend] {
     \label{fig:a}     
    \includegraphics[width=0.45\columnwidth]{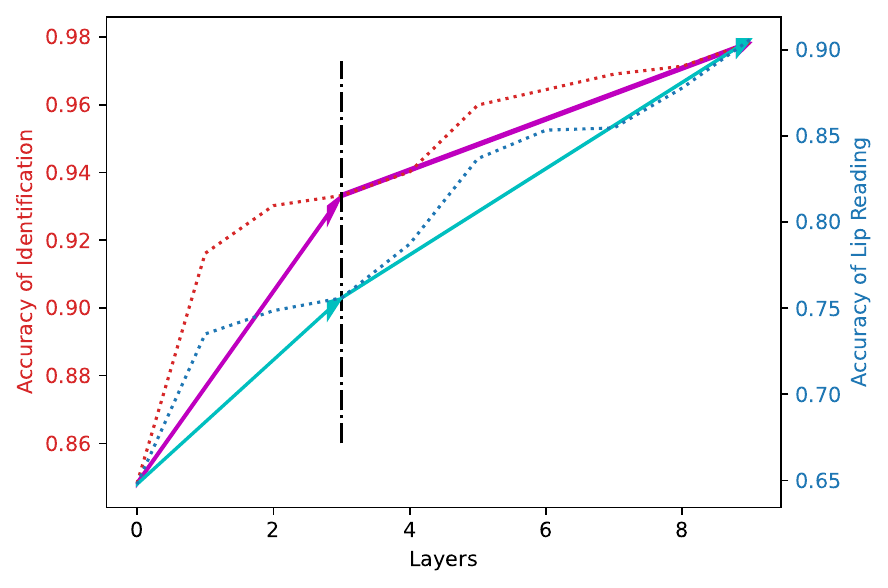}  
    }     
    \subfigure[Layer-by-layer Growth Trend] { 
    \label{fig:b}     
    \includegraphics[width=0.45\columnwidth]{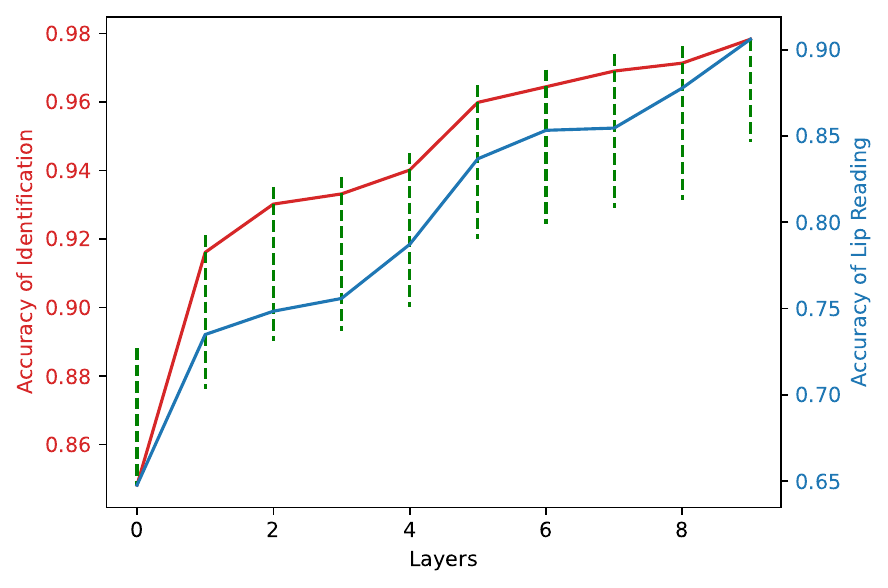}     
    }    
    \caption{Accuracy of Lip Reading and Identification Using the Output at Different Layers}
\label{fig:acc}
\vspace{-4ex}
\end{figure*}

In addition to the aforementioned observation, we also visualize features at different levels of a general lip reading model \cite{stafylakis2017combining, stafylakis2018pushing} and use them for speaker identification and lip reading tasks, respectively. As shown in Figure \ref{fig:acc}, the comparison of accuracy of shallow and deep layers, as well as the difference in performance improvement as the network depth increases, further confirms the above observation.

Based on the above analysis, we propose a novel speaker adaptive approach for lip reading by learning separable hidden unit contributions. 
We enhance the content-dependent features in shallow layers and suppress content-independent features in deeper layers, respectively. 
By adaptively enhancing content-dependent features with the speaker's features, we guide the model to focus on capturing content-dependent features which would be transferred to deeper layers to benefit the final lip reading task. 
Conversely, in deep layers where content-dependent features have obtained good representations, we introduce the speaker's feature to adaptively suppress the content-independent features, allowing the model to rely more on speech content-dependent features for robust lip reading.
%

Our contributions can be summarized as follows: 
1) We present new observations and analyses on lip-reading tasks for unseen speakers, focusing on two aspects: the distinction between speaker characteristics and speech content features at shallow and deep levels, and the dynamic role of speaker characteristics in recognizing different words.
2) We propose a new speaker adaptive lip reading method by taking advantage of the speaker's characteristics to learn for adaptively enhancing and suppressing separable hidden unit contributions for robust lip reading. 
3) For evaluation of unseen speaker's tasks in extreme settings, we release a new lip reading benchmark, CAS-VSR-S68, which involves only a few speakers but a diversified speech content range. Experiments on both the public dataset and our new benchmark show the advantages of our method.
\vspace{-1em}
\section{Related Work}
\vspace{-1em}
\par
\noindent\textbf{Lip Reading.} With the flourishing development of deep learning \cite{krizhevsky2017imagenet,hinton2012deep}, remarkable progress has been made in lip reading technology. At present, most lip reading models rely on the end-to-end deep learning framework, which can be broadly divided into visual feature extraction front-ends and language content decoding back-ends\cite{oghbaie2021advances}. Early researchers endeavored to enhance the visual feature and language feature modeling capabilities by modifying the neural network structure\cite{assael2016lipnet,stafylakis2017combining,stafylakis2018pushing,afouras2018deep} or extracting knowledge from other modalities, such as audio\cite{li2019improving,afouras2020asr} and deformation flow\citep{xiao2020deformation}.
Other researchers have created extensive audio-visual datasets from online videos to train models applicable to real-world scenarios\cite{yang2019lrw,Chung17,afouras2018deep,afouras2018lrs3,Chung16,chang2023conformers}. Recently, researchers have become increasingly interested in utilizing massive amounts of unlabeled data for self-supervised pre-training \cite{shi2022robust,lian2023av,haliassos2022jointly,baevski2022data2vec,baevski2022efficient}.
\par 
However, similar to automatic speech recognition (ASR), lip reading models always encounter performance degradation when dealing with unseen speakers \cite{assael2016lipnet, zhao2021unified}. The emerging topic of speaker adaptation in lip reading has gradually attracted attention in recent years, and some notable works have been conducted in this area \citep{kandala_speaker_2019,kim2022speaker, kim2023prompt,yang_speaker-independent_2020,zhang2021speaker,kandala2019speaker}. In comparison to their methods, we designed a new speaker adaptive method from a novel perspective, based on the analysis of the speaker's characteristics and their effect on lip reading. Our method achieved superior performance, both with and without extra adaptation data.
\par
\noindent\textbf{Speaker Adaptation.} A prevalent issue in audio-based Automatic Speech Recognition (ASR) is the significant degradation of recognition performance when testing conditions differ from the training conditions \cite{bell2020adaptation}. Speaker adaptation methods aim to address this problem. Some of these methods aim to extract more representative feature embeddings that capture speaker characteristics, including i-vectors \cite{saon2013speaker}, d-vectors \cite{variani2014deep}, x-vectors \cite{snyder2018x}, r-vectors \cite{khokhlov2019r}, and l-vectors \cite{meng2020vector}. Regularization-like approaches, such as meta-learning \cite{klejch_learning_2018, klejch_speaker_2019} or limiting the distance between the adaptation model and the speaker-independent model \cite{liao2013speaker, meng2019adversarial}, are also commonly used for speaker adaptation to mitigate overfitting. In some cases, data augmentation methods are used to alleviate data imbalance issues in datasets\cite{hosseini2018augmented}. 
\par

LHUC (Learning Hidden Unit Contribution) is a speaker adaptation method in which a separate linear layer is added to a pre-trained neural network to adjust the contribution of the hidden units for a specific speaker\cite{swietojanski2014learning}. This method has been successful in adapting models to different speakers in ASR tasks and has been adopted in many works\cite{xie2022variational,geng2022speaker,jin_personalized_2022,geng2022fly,jin2023adversarial,xie2019blhuc,swietojanski2016sat}.
Inspired by the LHUC approach, we have taken into account the differences between ASR and lip reading tasks, and developed a novel speaker adaptive method for lip reading based on our new observation and analysis of the speaker's characteristics and their effect on the lip reading task. 
%

\vspace{-1em}
\section{Our Proposed Method}
\vspace{-1em}
Based on the previous observation and analysis, we propose a speaker-adaptive method that utilizes speaker-dependent features to enhance and suppress features at shallow and deep layers of the lip reading network, respectively.
The overall architecture includes four modules, as shown in Figure \ref{fig:overview}: the speaker verification module, the feature enhancement module, the feature suppression module, and the lip reading module. 
The speaker verification module, represented by the yellow box on the left, aims to learn the speaker's characteristics. These characteristics are then passed to the feature enhancement module (lower middle green box) and the feature suppression module (upper middle pink box) to respectively generate speaker-dependent enhancement and suppression weights. These weights are then applied to the shallow and deep layers of the lip reading module (blue box on the left) respectively to adaptively enhance content-dependent features and suppress content-independent features. 
\vspace{-3ex}
\subsection{Model Architecture}
\vspace{-1ex}
\label{Architecture}
\begin{figure*}
\begin{center}
   \includegraphics[width=0.95\columnwidth]{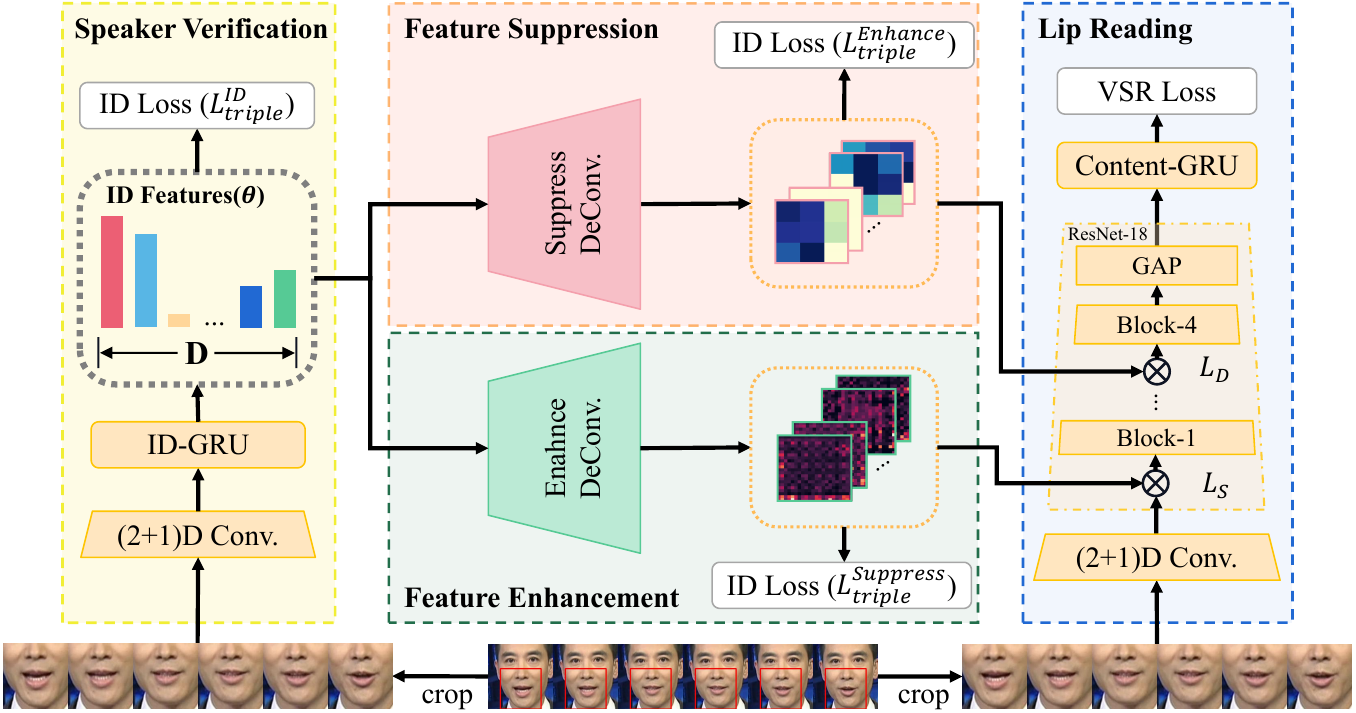}  
   \caption{The Overall Architecture of Our Proposed Method.}
   \label{fig:overview}
\end{center}
\vspace{-6ex}
\end{figure*}
We will introduce the four modules of our method in this part respectively.

\noindent\textbf{Speaker Verification Module (Yellow Box).} To learn speaker-dependent features for lip reading, we use speaker verification as a monitoring and guiding task during our model's learning process. 
Traditional face recognition models are not capable of modeling dynamic speaker-dependent features over time, such as speaking style, which can provide helpful auxiliary information for lip reading. Therefore, we propose a speaker verification module that crops each frame in the talking face video to the lip region and uses a (2+1)D convolutional network to model the spatial and temporal information. Then the speaker-dependent feature $\theta$ is generated with a single layer of GRU which is capable of modeling the temporal characteristics of the speaker. The obtained speaker-dependent feature finally serves as input to the subsequent feature enhancement and suppression modules.

\noindent\textbf{Lip Reading Module (Blue Box).} We adopt a popular lip reading network\cite{feng2020learn} as the backbone, which consists of a frontend network and a backend network. The frontend network includes a multi-layer (2+1)D convolutional neural network and a 2D ResNet-18, which models local temporal and spatial information in the sequence and aims to extract fine-grained visual features at each time step. The feature enhancement and suppression modules separately generate speaker adaptive weights, which are element-wise multiplied with the shallow and deep layer features of ResNet-18, respectively. The backend network consists of a 3-layer bidirectional GRU that models global temporal and linguistic information and generates content-dependent features for lip reading.

\noindent\textbf{Feature Enhancement/Suppression Module (Green/Pink Box).} 
The feature enhancement module is used for shallow layers where the speech content features are less prominent, in order to adaptively take advantage of the speaker's characteristics to enhance the speech content features. 
It generates enhancement weights based on the speaker's features $\theta$ to guide the model to focus on content-dependent features. 
Then these weights are applied to the shallow layers $L_S$ (the first layer of the 2D ResNet-18) in the lip reading module. 
Specifically, we introduce an enhancement deconvolutional network to upsample the speaker's features $\theta$ to the final enhancements weights of the same size and channel number as the target shallow layer $L_S$ of the lip reading module. The enhancement weights are generated with the activation function $\sigma_{enhance}(h)=1+|\mathrm{LeakyReLU}(h)|$ after enhancement deconvolutional layers. 
As a result, the output values are bounded to the range of [1,+$\infty$), ensuring that content features that are originally weak in the shallow layers of the lip reading module are only enhanced adaptively, rather than being suppressed. This approach is beneficial for achieving robust lip reading features to feed subsequent stages. 

For deep layers where the speech content features have been obtained well, the feature suppression module is introduced to take the speaker's characteristics to generate suppression weights to adaptively suppress irrelevant noise for robust lip reading. 
This module also takes the speaker's feature $\theta$ as input but generates the suppression weights through another deconvolutional network as shown in Figure \ref{fig:overview}. These weights are then applied to the deep layers $L_D$(14th layer of the 2D ResNet-18) of the lip reading module. We apply an activation function of $\sigma_{suppress}(h)=1-|\mathrm{tanh}(h)|$ after the suppression deconvolutional layers to obtain values within the range of (0,1]. 
The operation of adaptive suppression aims to eliminate only the irrelevant noise, rather than content-dependent features, to minimize the risk of model overfitting. 

\vspace{-2ex}
\subsection{Optimization}
\vspace{-1ex}
Four losses of the architecture are shown in the white solid box in Figure \ref{fig:overview}.

\noindent\textbf{ID Loss for Speaker Verification}. To ensure the effective extraction of speaker characteristics by the speaker verification module, we introduce the triplet loss \cite{schroff2015facenet}. This helps the model map ID features to a generalizable feature space while avoiding the use of an excessively large linear layer.
Specifically, during the sample selection process, we choose a positive sample $A'_{ID}$ of the same speaker as the anchor video $A_{ID}$, and a negative sample $B_{ID}$ of a different speaker. Then, we simultaneously input the three samples into the speaker verification module to extract the speaker's features $\theta$,  denoted as $\theta_{A_{ID}}$, $\theta_{A'_{ID}}$, and $\theta_{B_{ID}}$, respectively. The optimization of this loss function is as:
\vspace{-1ex}
\begin{equation}
L^{ID}_{triple} (\theta_{A_{ID}},\theta_{A'_{ID}},\theta_{B_{ID}}) = \max(d(\theta_{A_{ID}}-\theta_{A'_{ID}})-d(\theta_{A_{ID}}-\theta_{B_{ID}})+\alpha_t,0)
\vspace{-1ex}
\end{equation}
where $\alpha_t$ is a margin between positive and negative samples of ID features, and the Euclidean distance is used as the distance function, i.e., $d(A,B)=||A-B||^2$.

\noindent\textbf{ID Loss for Feature Enhancement/Suppression Module.} To avoid the model collapsing to trivial solutions, where all generated enhancement and suppression weights are equal to 1, we add speaker-level supervision to the generated weights to ensure that they exhibit differences across speakers. Similar to the loss function of the speaker verification module, we first input the speaker's feature $\theta$ extracted by the speaker verification module into the feature enhancement network to obtain the corresponding enhancement weights $\theta^{Enhance}$. The optimized loss can be represented as
\vspace{-1ex}
\begin{equation}
\begin{aligned}
L^{Enhance}_{triple}(A_{ID},{A'}_{ID},B_{ID})=\max(d(\theta^{Enhance}_{A_{ID}}-\theta^{Enhance}_{{A'}_{ID}})-d(\theta^{Enhance}_{A_{ID}}-\theta^{Enhance}_{B_{ID}})+\alpha_E,0)
\end{aligned}
\vspace{-1ex}
\end{equation}
where $\alpha_E$ represents the margin that distinguishes the weights generated for positive and negative samples in the enhancement process.

For the suppression weights, we use a similar approach to calculate the optimization loss:
\vspace{-1ex}
\begin{equation}
\begin{aligned}
L^{Suppress}_{triple}(A_{ID},{A'}_{ID},B_{ID})=\max(d(\theta^{Suppress}_{A_{ID}}-\theta^{Suppress}_{{A'}_{ID}})-d(\theta^{Suppress}_{A_{ID}}-\theta^{Suppress}_{B_{ID}})+\alpha_S,0)
\end{aligned}
\vspace{-1ex}
\end{equation}
where $\alpha_S$ represents the margin that distinguishes the weights generated for positive and negative samples in the suppression process.
The choice of $\alpha_S$ was determined based on the proportion of the mean and variance of the two types of weights since the generated enhancement and suppression weights have corresponding value ranges. We used the Euclidean distance as the distance measure d.

\noindent\textbf{VSR Loss.} We have carried out evaluations at both the word-level and sentence-level for lip-reading tasks. For word-level lip reading task, we use the cross-entropy loss for training, represented as
\begin{equation}
    \vspace{-2ex}
L^{VSR}_{CE} = -\frac{1}{N}\sum_{i=1}^{N}\sum_{j=1}^{|V|} y_{i,j}\log(\hat{y}_{i,j})
\end{equation}
where $N$ is the total number of training samples, $|V|$ is the vocabulary size, $y_{i,j}$ is the ground truth label, and $\hat{y}_{i,j}$ is the predicted probability of the $j$-th word for the $i$-th sample.
For sentence-level lip reading, we used the CTC loss \cite{graves2006connectionist} for optimization, represented as 
\begin{equation}
    \vspace{-1ex}
    L^{VSR}_{CTC} = -\log\sum_{\pi\in\mathcal{B}^{-1}(y)} p(\pi|x)
    \vspace{-1ex}
\end{equation}
where $\mathcal{B}^{-1}(y)$ denotes the set of all possible alignments of the label sequence $y$ and $p(\pi|x)$ is the probability of the alignment $\pi$ given the predicted input sequence $x$.

\vspace{-3ex}
\section{Experiments}
\vspace{-2ex}
\subsection{Datasets}
\vspace{-1ex}
\noindent \textbf{GRID}\cite{cooke2006audio} corpus comprises 1000 utterances spoken by 33 speakers, with both audio and visual data captured simultaneously. We use the widely-adopted split of unseen speakers as in LipNet\cite{assael2016lipnet}, where the 1st, 2nd, 20th, and 22nd speakers serve as the test set while the remaining speakers are used for training. When evaluating in the setting of providing adaptation data, we randomly divide the four speakers of the original test set of GRID into adaptation and evaluation sets for each speaker\cite{kim2022speaker}.

\noindent \textbf{LRW-ID}\cite{kim2022speaker} is a redivision of the LRW dataset\cite{Chung17}. LRW is a large-scale dataset for lip reading, which consists of 500 different words spoken by more than 10 thousand speakers in the wild. Kim et al. \cite{kim2022speaker} use face recognition technology to annotate the identity of speakers and redivide the original training and testing sets to evaluate speaker-adaptive lip reading methods. Specifically, LRW-ID selects data of 20 speakers as validation (adaptation) and test sets, and the remaining data is used for training. We adopt the same division for evaluation.

\noindent \textbf{CAS-VSR-S68} is a new dataset for evaluation in the extreme setting of unseen speakers' lip reading. The data was collected from news broadcast programs aired between 2009 and 2019, covering a wide range of topics and speakers. The video clips and corresponding text annotations were extracted from the broadcast clips of the hosts, with a resolution of 256*256 pixels and sentence lengths varying from one character to over 20 characters. The dataset has a total duration of approximately 68 hours and includes over 3800 commonly used Chinese characters, making it a challenging dataset for lip reading tasks. The dataset contains data from 11 hosts, with 10 hosts' data used for training and the remaining host's data randomly split into adaptation and testing sets.

The detailed settings for each dataset are given in the supplementary materials.
\vspace{-3ex}
\subsection{Results}
\vspace{-2ex}
\subsubsection{Qualitative Analysis}
\vspace{-1ex}
\begin{figure*}
\begin{center}
   \includegraphics[width=0.95\columnwidth]{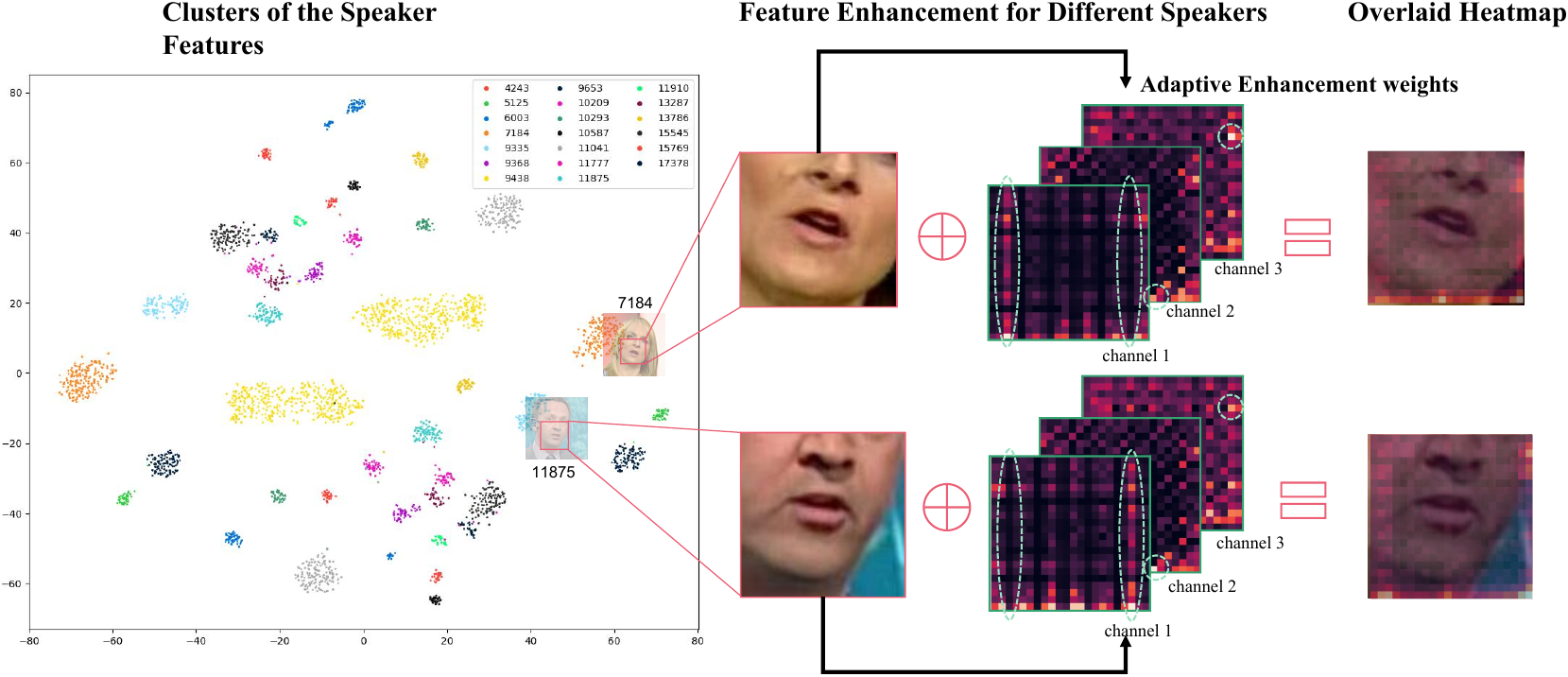}  
   \caption{Visualization of the Generated Enhancement Weights}
   \label{fig:FM}
\end{center}
\vspace{-3ex}
\end{figure*}
\textbf{Discriminative Speaker's Features.} Figure \ref{fig:FM} shows the visualization of speaker features $\theta$ for each speaker in the LRW-ID dataset using t-SNE dimensionality reduction.We performed clustering on the reduced speakers' feature $\theta$, and each cluster is generally gathered together, effectively capturing the unique characteristics of each speaker.

\noindent\textbf{Adaptive Enhancement Weights.} 
We produce heat map visualizations to represent the enhancement weights for randomly selected samples, as shown in the middle of Figure \ref{fig:FM}. The enhancement weights for the same channel display significant differences across different speakers. For instance, the green circle in the figure highlights the significant contrast between the two randomly sampled speakers. The same region may be significantly enhanced for one speaker, while the enhancement in the same region for another speaker may not be obvious.
 

\noindent\textbf{Visual Analysis of Enhanced Features.}
We compute the average of the feature enhancement weights along the channel dimension and overlay them on the corresponding speaker's input image with some transparency. This enables us to analyze which part of the information on the face would the feature enhancement weights intend to focus on. As displayed on the right-hand side of Figure \ref{fig:FM}, the feature enhancement weights are noticeably concentrated in the regions close to the image edges. 
This is because lip reading networks usually focus on the central lip region of the image, neglecting helpful information from other regions such as the chin, cheeks, and nose. In fact, it has been shown that these areas contain speech-related information indeed\cite{zhang2020can}. Our proposed method is able to enhance the model's attention to areas beyond the lips, therefore enabling more effective recognition of the speaker's speech content.

The suppression weight also exhibits a similar property to the enhancement weight, and detailed analyses are provided in the supplementary materials.
\vspace{-2ex}
\subsubsection{Quantitative Analysis}
\begin{table}[t]
\begin{minipage}[t]{.60\linewidth}
  \centering
  \caption{Ablation Study of Loss Functions}
  \label{tab:2}
  \begin{tabular}{ccccc}
    \toprule
    \textbf{Method} & $\bf{L^{ID}_{triple}}$ & $\bf{L^{Enh}_{triple} \& L^{Sup}_{triple}}$ & $\bf{L^{VSR}_{CE}}$ & \textbf{Acc(\%)}\\
    \midrule
    Baseline & - & - & \checkmark & 87.25 \\
    \midrule
    \multirow{4}{*}{\centering Ours} & \ding{53} & \ding{53} & \checkmark & 87.73 \\
    & \checkmark  & \ding{53} & \checkmark & 87.74 \\
    & \ding{53} & \checkmark & \checkmark & 87.75 \\
    & \checkmark & \checkmark & \checkmark & \textbf{87.91} \\
    \bottomrule
  \end{tabular}
\end{minipage}
\hfill
\begin{minipage}[t]{.4\linewidth}
  \centering
  \caption{Ablation Study of Modules}
  \label{tab:3}
  \begin{tabular}{cc}
    \toprule
    \textbf{Method} & \textbf{Acc(\%)}\\
    \midrule
    Baseline & 87.25 \\
    Enhance Only & 87.83 \\
    Suppress Only & 87.81 \\
    Proposed & \textbf{87.91} \\
    \bottomrule
  \end{tabular}
\end{minipage}
\end{table}
\begin{table}[bt]
\centering
\caption{Performance on LRW-ID with Reduced Data and Speaker Diversity}
\label{tab:1}
\begin{tabular}{c|cc|cc|cc}
\toprule
\-&\textbf{\makecell{Sample\\size}} & \textbf{\makecell{Number of\\speakers}} & \textbf{\makecell{Acc(\%)\\Baseline}} & \textbf{\makecell{Acc(\%)\\Ours}} & \textbf{\makecell{Perf.Drop\\Baseline}} & \textbf{\makecell{Perf.Drop\\Ours}} \\
\midrule
\textbf{a} & 480378 & 17560 & 87.25 & 87.91 & - & - \\
\midrule
\textbf{b} & 383788 & 17551 & 85.26 & 86.11 & $\downarrow$ 2.28\% & $\downarrow$ \textbf{2.05\%} \\
\textbf{c} & 388577 & 1047 & 85.4 & 86.36 & $\downarrow$ 2.12\% & $\downarrow$ \textbf{1.76\%} \\
\midrule
\textbf{d} & 386681 & 1000 & 85.2 & 86.27 & $\downarrow$ 2.35\% & $\downarrow$ \textbf{1.87\%} \\
\textbf{e} & 354807 & 500 & 84.22 & 84.88 & $\downarrow$ 3.47\% & $\downarrow$ \textbf{3.45\%} \\
\textbf{f} & 298369 & 200 & 81.37 & 82.7 & $\downarrow$ 6.74\% & $\downarrow$ \textbf{5.93\%} \\
\textbf{g} & 246108 & 100 & 77.46 & 78.5 & $\downarrow$ 11.22\% & $\downarrow$ \textbf{10.70\%} \\
\bottomrule
\end{tabular}
\end{table}

\begin{table}[bt]
\vspace{-2ex}
\centering
\caption{Comparison with Other Methods on LRW-ID and CAS-VSR-S68}
    \resizebox{\textwidth}{!}{\begin{tabular}{cccccc|cc}
    \toprule
    \multirow{2}{*}{\makecell{Adapt\\min.}} & \multicolumn{5}{c}{\textbf{LRW-ID(ACC\%)}} & \multicolumn{2}{c}{\textbf{CAS-VSR-S68(CER\%)}}  \\
    \cmidrule{2-6} \cmidrule{7-8}
     & \makecell{User-\\padding\cite{kim2022speaker}} & \makecell{Prompt\\Tuning\cite{kim2023prompt}} & \makecell{DCTCN\\ \cite{ma2022training}}& Baseline & \makecell{Proposed\\Method} & Baseline & \makecell{Proposed\\Method} \\
    \midrule
    0 & 85.85 & 87.54 & 86.75 & 87.25 & \bf{87.91} & 19.61 & 19.37 \\
    1 & 87.06 & 88.53 & -     & 88.52 & \bf{89.21} & \underline{21.53} & \underline{20.69} \\
    3 & 87.61 & 89.45 & -     & 89.48 & \bf{89.88} & 18.65 & 18.55 \\
    5 & 87.91 & 89.99 & -     & 89.96 & \bf{90.45} & 17.55 & \bf{16.72}\\
    \bottomrule
    \end{tabular}}%
\label{tab:comparison}%
\vspace{-3ex}
\end{table}
\noindent\textbf{Ablation Study.} The ablation study in Table \ref{tab:2} shows the effectiveness of each loss in our method. 
$L^{ID}_{triple}$ constrains the speaker verification module to extract each speaker's own features correctly. $L^{Enhance}_{triple}$ and $L^{Suppress}_{triple}$ are introduced to prevent the collapse of the generated weights. Therefore, removing either loss results in consistent performance degradation.

The ablation experiments in Table \ref{tab:3} show the effectiveness of each module. The enhance and suppress modules improve the model's ability to use speaker-dependent information to handle unseen speakers. When combined, they further improve the model's ability to handle unseen speakers, demonstrating the effectiveness of our method.

\noindent\textbf{Reduce Speaker Diversity.} Usually, the lip reading model's performance on unseen speakers would decline as the sample size decreases or speaker diversity reduces for training. As shown in Table \ref{tab:1}, our proposed method effectively alleviates this degradation in performance caused by both the reduction in speaker diversity and sample size in the training set. 
Notably, our method demonstrates significant mitigation of performance loss in scenarios where the sample size is relatively close but the number of speakers is small, as observed in experiments \textbf{b} and \textbf{c}, indicating that our method utilizes speaker information sufficiently. 
Furthermore, experiments from \textbf{d} to \textbf{g} demonstrate that our model generally mitigates performance degradation as the speaker diversity decreases, even in scenarios with only 100 speakers remaining.
\begin{figure*}[hbt]
\begin{center}
   \includegraphics[width=0.95\columnwidth]{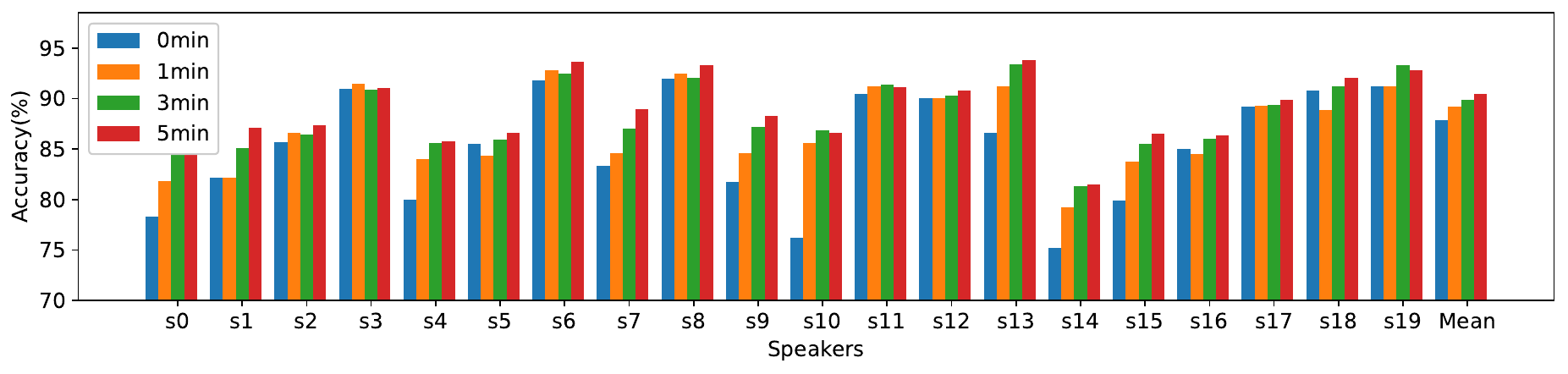} 
   \caption{Adaptation Rresult Using Different Amount of Adaptation Data on LRW-ID}
   \label{fig:adp}
   \vspace{-4ex}
\end{center}
\end{figure*}
\begin{table}[hbt]
\centering

\begin{minipage}[t]{.48\textwidth}
  \centering
  \resizebox{\linewidth}{!}{ 
  \begin{threeparttable}
    \caption{Comparison Results on GRID with Other Methods without any Adaptation Data}
    \label{tab:4}
    \begin{tabular}{l>{\centering\arraybackslash}m{3.5cm}}
      \toprule
      \textbf{Method} & \textbf{WER(\%)} \\
      \midrule
      WAS\cite{Chung17}\tnote{**}&14.6\\
      LipNet\cite{assael2016lipnet}\tnote{**}& 11.4 \\
      TM-seq2seq\cite{afouras2018deep}\tnote{**}&11.7\\
      User-padding\cite{kim2022speaker}  & 11.12 \\
      User-padding\cite{kim2022speaker}\tnote{*} & 7.2 \\
      Prompt Tuning\cite{kim2023prompt}  & 12.04 \\
      TVSR-Net\cite{yang_speaker-independent_2020}  & 9.1 \\
      DVSR-Net\cite{zhang2021speaker}  & 7.8 \\
      Visual i-vector\cite{kandala2019speaker}  & 7.3 \\
      \midrule
      Baseline (ours)  & 10.62 \\
      Proposed Method  & 9.59 \\
      Proposed Method\tnote{*} & \bf{6.99} \\
      \bottomrule
    \end{tabular}
    \begin{tablenotes}
      \footnotesize
      \item[*]Using our method in the manner as  \cite{kim2022speaker}  
      \item[**] Results reproduced by \cite{yang_speaker-independent_2020,zhang2021speaker} 
    \end{tablenotes}
  \end{threeparttable}
  }
\end{minipage}%
\hfill
\begin{minipage}[t]{.40\textwidth}
  \centering
  \resizebox{\linewidth}{!}{ 
  \begin{threeparttable}
    \caption{Adaptation Result on GRID Dataset}
    \label{tab:adaptation_results}
    \begin{tabular}{lcc}
      \toprule
      \textbf{Method} & \textbf{Adapt min.} & \textbf{WER(\%)} \\
      \midrule
      \multirow{4}{*}{\makecell{User\\Padding\cite{kim2022speaker}}} & 0 & 11.12 \\
      & 1 & 6.8 \\
      & 3 & 6.05 \\
      & 5 & 5.67 \\ 
      \midrule
      \multirow{4}{*}{\makecell{Prompt\\Tuning\cite{kim2023prompt}} }& 0 & 12.04 \\
      & 1 & 5.53 \\
      & 3 & 4.31 \\
      & 5 & 3.8 \\ 
      \midrule
      \multirow{4}{*}{\makecell{Proposed\\Method}} & 0 & 9.59 \\
      & 1 & 5.61 \\
      & 3 & 4.6 \\
      & 5 & \bf{3.59} \\
      \bottomrule
    \end{tabular}
  \end{threeparttable}
  }
\end{minipage}
\end{table}

\noindent \textbf{Without Adaptation Data.} 
Our proposed lip reading method performs well in scenarios where there is no adaptation data for unseen speakers.
From Table \ref{tab:comparison}, it can be seen that our proposed method not only achieves an overall improvement compared to the LRW-ID baseline but also outperforms the previous works. 
The comparison between the baseline and our proposed method in Table \ref{tab:4} demonstrates that even when the baseline performance is already relatively high, our method still achieves a performance improvement of about relativelt 9.70\% on GRID.
By applying the unsupervised speaker adaptation technique proposed in \cite{kim2022speaker}, our model achieves a significant improvement in WER, reducing it from 9.59\% to 6.99\%, which corresponds to an additional performance gain of approximately 27.1\%.
We also observed performance gains on the CAS-VSR-S68 dataset, which represents an extreme case characterized by limited speaker diversity and a broad scope of speech content. This is a challenging dataset, with only 1-minute short adaptation data, the performance decreased instead of increasing, which fully illustrates the extreme situation and challenges faced by this dataset. 
In summary, our method shows its ability on effective utilization of speaker-dependent information to improve the model's ability to recognize unseen speakers, regardless of the speaker diversity of the dataset, as observed in LRW-ID (20k+ speakers), GRID (29 speakers), and CAS-VSR-S68 (9 speakers).

\noindent\textbf{With Limited Adaptation Data.} 
In addition to the model's excellent performance in the absence of any adaptation data, our method can also effectively utilize limited speaker adaptation data (<5min) to significantly improve the model's performance, outperforming other methods. 
As shown in Figure \ref{fig:adp}, the model's performance generally increases as the amount of adaptation data increases. Combining with Tables \ref{tab:comparison} and \ref{tab:adaptation_results}, our method shows a significant advantage on the LRW-ID and GRID datasets, achieving an accuracy of 90.45\% and a WER of 3.59\% respectively with only about 5 minutes of training data. 
\section{Conclusion}
This paper proposes a novel speaker adaptive method for lip reading of unseen speakers, leveraging the speaker's own characteristics to learn separable hidden unit contributions. The approach outperforms existing methods on public popular datasets and our challenging dataset CAS-VSR-S68 with a few speakers but diverse speech content. Our method provides a new idea and solution for robust visual speech recognition.
\section{Acknowledgements}
This work is partially supported by National Natural Science Foundation of China (No. 62276247, 62076250). We thank Yuanhang Zhang for his contributions to the collection, download, and processing of the CAS-VSR-S68 dataset.
\bibliography{egbib}
\end{document}


\maketitle

In this supplementary material, we provide additional insights and analyses of our method for lip reading. Specifically, Section \hyperref[Introduction]{1} illustrates the distinction between speaker-dependent and content-dependent features extracted by lip reading models. Section \hyperref[sec:exp]{2} presents more experimental results, including more detailed quantitative results and qualitative visualizations, which further prove the effectiveness of our proposed approach.
\section{Illustration of the Features Extracted by Lip Reading Models}
\label{Introduction}
\par
\begin{figure*}[htbp]
\begin{center}
   \includegraphics[width=1\columnwidth]{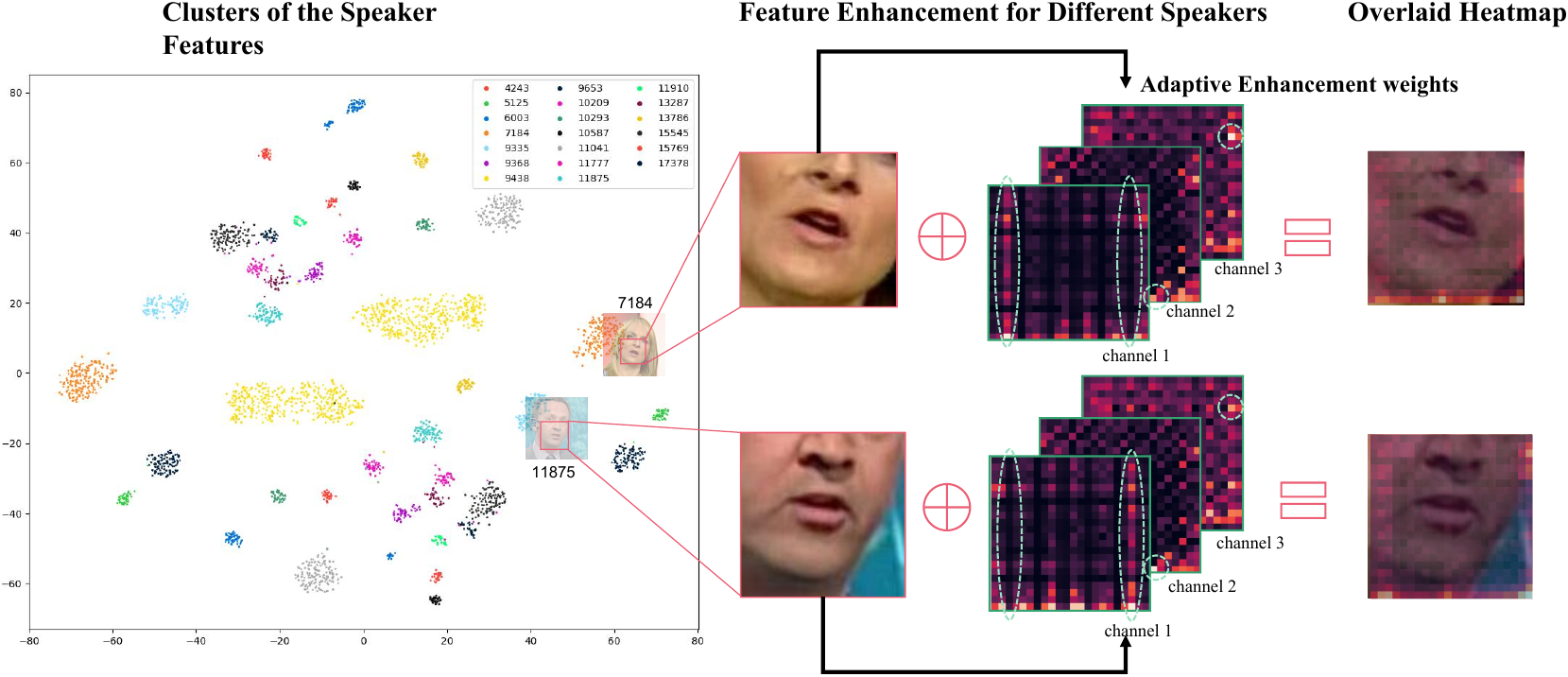}  
   \caption{Illustration of the relationships between the speaker-dependent and the content-dependent features.}
   \label{fig:feat}
\end{center}
\end{figure*}

Given any lip reading model, we can roughly divide the features extracted by this model into two types according to different criteria: speaker-dependent and speaker-independent features, or content-dependent and content-independent features. The features under these two criteria focus on expressing different properties when give a talking face video. We show these two types as the two ends with different colors in Figure \ref{fig:feat}.

\textbf{Speaker-dependent features} primarily capture the unique characteristics of the speaker and are always reflected by the speaker's static facial traits, such as mouth shape, skin texture, skin tone, beard, markings, and also a few dynamic traits corresponding with the speaker's speaking style. These features encode the speaker's individuality and remain relatively constant when speaking different words or utterances. They significantly contribute to the overall process of identifying and differentiating the speaker from other individuals.

\textbf{Content-dependent features} are closely related to the specific spoken content and mainly focus on the fine-grained spatio-temporal changes in the facial region, especially the lip region, during the speaking process. They are more sensitive to the specific words being pronounced, which provides the basis for the lip reading task.

As discussed in the main submission, there exists an interesting phenomenon regarding the performance of shallow-layer and deep-layer features for speaker identification and lip reading tasks. The accuracy of speaker identification using shallow-layer features is already high, and as the layers go deeper in the network, the accuracy of speaker identification experiences a rapid increase. However, when utilizing the same shallow-layer features for lip reading, the recognition accuracy is relatively low, and the rate of improvement in accuracy is much slower compared to speaker identification. Intriguingly, the rate of increase in lip reading task accuracy is generally higher than that of speaker classification accuracy in the deep layers of the network.
This phenomenon highlights the distinctive nature of our method to learn separable hidden unit contributions for shallow and deep layers respectively.
%
\section{More Detailed Experiments}
\label{sec:exp}
\subsection{Training Details}

We employ a three-step training approach to learn our model as shown in Figure 2 in the main submission, to learn the contradictory targets of the enhancement and suppression module.
Firstly, we train the left speaker verification module with $L^{ID}_{triple}$ and the right lip reading modules with $L^{VSR}_{CE}$ separately. 
Then, we introduce the feature enhancement module together with  the learned speaker verification module and the lip reading module to continue the training process.  
Finally, we freeze the feature enhancement module and the speaker verification module to introduce the suppression module to continue training until convergence. 
\subsection{Experimental Setup}

\noindent\textbf{LRW-ID:}
We utilized the Adam\cite{kingma2014adam} optimizer with a maximum learning rate of $8.125 \times 10^{-4}$ and a batch size of 130. The input size was $29 \times 96 \times 96$ (T, W, H), where T represents the number of frames. Data augmentation techniques included horizontal flipping and random cropping to $88 \times 88$. 
The fine-tuning phase with adaptation data involved the Adam optimizer with a maximum learning rate of $6.25 \times 10^{-5}$ and a batch size of 200. 

\noindent\textbf{GRID:}
The Adam optimizer with a maximum learning rate of $1.5 \times 10^{-4}$ and a batch size of 32 was employed. The input size was set to $29 \times 96 \times 96$ (T, W, H), and data augmentation techniques included horizontal flipping and random cropping to $88 \times 88$. Fine-tuning with adaptation data was performed using the Adam optimizer with a maximum learning rate of $7.25 \times 10^{-5}$ and a batch size of 32. 

\noindent\textbf{CAS-VSR-S68:}
The Adam optimizer with a maximum learning rate of $8.125 \times 10^{-4}$ and a dynamic batch strategy was used. The maximum input frame count was set to 300, and the input size was $T \times 96 \times 96$ (T, W, H). Data augmentation included horizontal flipping with a 0.5 probability. 
For fine-tuning with adaptation data, the Adam optimizer with a maximum learning rate of $6.25 \times 10^{-5}$ and a batch size of 1 was used. 
Similar to the previous datasets, 1 minute, 3 minutes, and 5 minutes of data from the adaptation set were randomly selected for full model fine-tuning. 
\subsection{Results}
\subsubsection{More Quantitative Results}

\begin{table}[ht]
\centering
\caption{Comparison on GRID with Other Methods without Any Adaptation Data}
\label{tab:4}
 \begin{threeparttable}
\begin{tabular}{lccccc}

\toprule
\multirow{2}{*}{\textbf{Method}} & \multicolumn{4}{c}{\textbf{Test Speaker}} & \multirow{2}{*}{\textbf{Mean(WER)}} \\ \cmidrule(lr){2-5}
 & S1 & S2 & S20 & S22 &\\
\midrule
LipNet (reproduce)\cite{assael2016lipnet} & 22.13 & 10.42 & 11.83 & 6.73 & 13.6 \\
User-padding\cite{kim2022speaker} & 17.04 & 9.02 & 10.33 & 8.13 & 11.12 \\
User-padding\cite{kim2022speaker}\tnote{*} & - & - & - & - & 7.2 \\
Prompt Tuning\cite{kim2023prompt} & 16.4 & 9.42 & 11.23 & 11.57 & 12.04 \\
DVSR-Net\cite{yang_speaker-independent_2020} & - & - & - & - & 9.1 \\
TVSR-Net\cite{zhang2021speaker} & - & - & - & - & 7.8 \\
Visual i-vector\cite{kandala2019speaker}  & - & - & - & - & 7.3 \\
Baseline (ours) & 19.60 & 10.96 & 7.26 & 4.65 & 10.62 \\
Proposed Method & 17.96& 9.20 & 6.46 & 4.72 & 9.59 \\
Proposed Method\tnote{*} & 13.01 & 5.63 & 5.86 & 3.45 & \bf{6.99} \\

\bottomrule

\end{tabular}
    \begin{tablenotes}
      \footnotesize
      \item[*]Test in the manner as  \cite{kim2022speaker} 
    \end{tablenotes}
      \end{threeparttable}

\end{table}

\begin{table}[ht]
\centering
\caption{Using Different Quantities of Adaptation Data on GRID}
\label{tab:adaptation_results}
\begin{tabular}{lcccccc}
\toprule
\multirow{2}{*}{\textbf{Method}} & \multirow{2}{*}{\textbf{Adapt min.}} & \multicolumn{4}{c}{\textbf{Test Speaker}} & \multirow{2}{*}{\textbf{Mean(WER)}} \\ \cmidrule(lr){3-6}
 &  & S1 & S2 & S20 & S22 &  \\ \midrule
\multirow{4}{*}{User Padding\cite{kim2022speaker} } & 0 & 17.04 & 9.02 & 10.33 & 8.13 & 11.12 \\
 & 1 & 10.65 & 4.2 & 7.77 & 4.59 & 6.8 \\
 & 3 & 9.35 & 3.75 & 6.88 & 4.27 & 6.05 \\
 & 5 & 8.78 & 3.45 & 6.49 & 3.99 & 5.67 \\ \cmidrule(lr){1-7}
\multirow{4}{*}{Prompt Tuning\cite{kim2023prompt} } & 0 & 16.40 & 9.42 & 11.23 & 11.57 & 12.04 \\
 & 1 & 7.91 & 3.81 & 6.07 & 4.43 & 5.53 \\
 & 3 & 6.43 & \bf{2.14} & 5.63 & 3.07 & 4.31 \\
 & 5 & 5.08 & 2.24 & 5.13 & 2.8 & 3.8 \\ \cmidrule(lr){1-7}
\multirow{4}{*}{Proposed Method} & 0 & 17.96 & 9.20 & 6.46 & 4.72 & 9.59 \\
 & 1 & 9.22 & 4.53 & 5.59 & 3.11 & 5.61 \\
 & 3 & 6.52 & 3.2 & 6.12 & \bf{2.54} & 4.60 \\
 & 5 & \bf{4.78} & 2.53 & \bf{4.38} & 2.68 & \bf{3.59} \\ \bottomrule
\end{tabular}
\end{table}

\noindent\textbf{Detailed Results on GRID.} 
In our main submission, we evaluated the effectiveness of our method on the GRID dataset by measuring the Word Error Rate (WER), both with and without adaptation data. In this section, we provide a detailed analysis of the experimental results for each speaker and compare our method with other approaches.

Table \ref{tab:4} clearly shows that our method consistently outperforms the comparison methods in terms of WER across all speakers, even in the absence of adaptation data. Additionally, our method exhibits a notable overall performance improvement compared to the baseline. We specifically observe significant improvements for speakers S1 and S2, who initially had higher WER.
Moreover, Table \ref{tab:adaptation_results} demonstrates consistent improvements achieved across different speakers when adaptation data is available. Remarkably, with a mere 5 minutes of adaptation data, the WER for Speaker 1 significantly decreases from 17.96 to 4.78, representing an impressive reduction of approximately 73.4\%.
\begin{table}[t]
\centering
\caption{Character Error Rate (CER) on CAS-VSR-S68 with Adaptation Data}
\label{tab:xwlb}
\begin{tabular}{ccc}
\toprule
Adapt min & Baseline & Proposed Method \\
\midrule
0 & 44.93 & 43.24 \\
1 & 38.63 & 37.38 \\
3 & 36.37 & 35.64 \\
5 & 33.79 & \bf{33.17} \\
\bottomrule
\end{tabular}
\end{table}
\begin{figure*}[t]
\begin{center}
   \includegraphics[width=1\columnwidth]{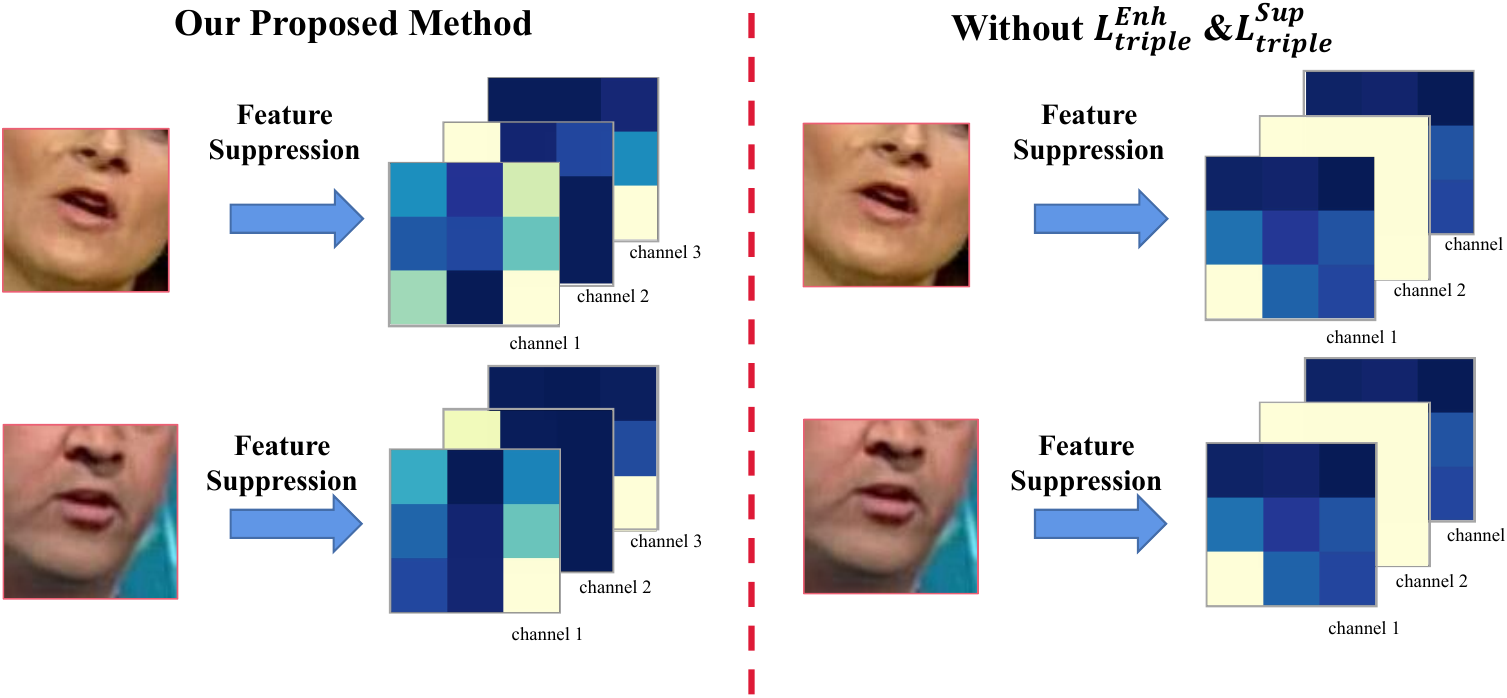}  
   \caption{Visualization of Features Generated by Suppression Module}
   \label{fig:supp}
\end{center}
\end{figure*}

\noindent\textbf{Detailed Results on CAS-VSR-S68.} 
To further validate the effectiveness of our method, we conducted additional experiments using a different test set. In the main submission, we utilized the data of a male news anchor, Gang Qiang, as the test set. In addition to that, we also evaluated our method on a separate test set consisting of data from a female news anchor, Li Ruiying, as shown in Table \ref{tab:xwlb}.

It is worth noting that due to the limited amount of female data in the training set, the baseline performance on the female news anchor was relatively lower compared to the male news anchor. However, our proposed method consistently outperformed the baseline across different adaptation settings, demonstrating its effectiveness in improving lip reading performance.
Furthermore, we did not observe the unusual performance decrease when using only 1-minute short adaptation data, as mentioned in the main submission. This suggests that the extreme situation and challenges faced by the CAS-VSR-S68 dataset may have different underlying factors that require further investigation in future research.

Overall, the additional experiments provide further evidence of the effectiveness of our method in improving lip reading performance. They highlight the importance of considering speaker diversity and addressing the challenges posed by different speakers in the dataset.

\begin{figure*}[ht]
\begin{center}
   \includegraphics[width=1\columnwidth]{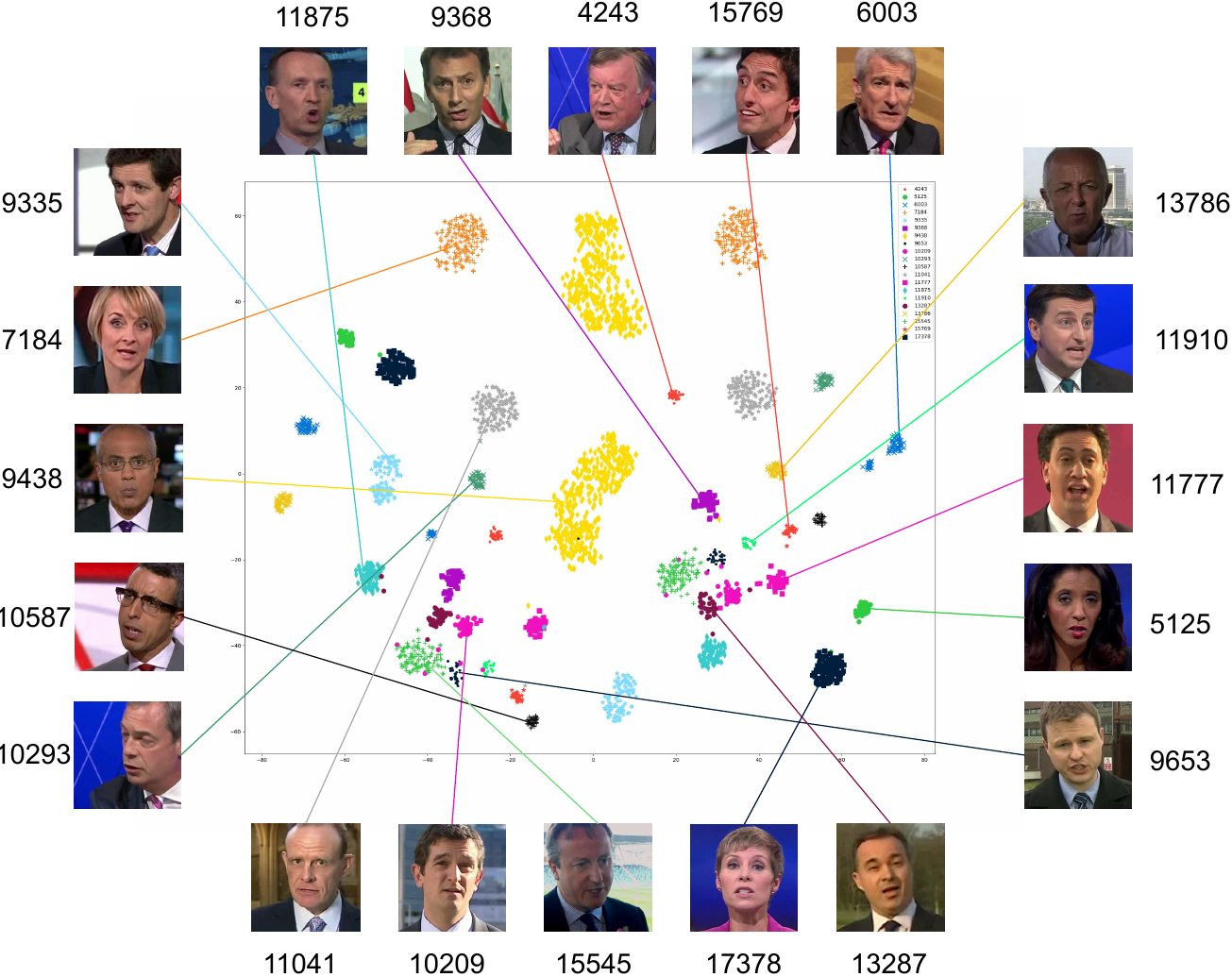}  
   \caption{Clustering Visualization of the Learned ID Features Obtained through t-SNE Dimensionality Reduction on the LRW-ID Dataset. \\
   Due to the similarity in colors of the figures presented in the main text, which made them less distinguishable, we have made modifications to the legend. We performed t-SNE dimensionality reduction on the same set of samples to obtain a clearer visualization. The clustering results in the revised figure show some shifting compared to the clusters mentioned in the original text.}
   \label{fig:tsne}
\end{center}
\end{figure*}

\subsubsection{More Qualitative Results}
\noindent\textbf{Further Analysis of Alblation Study.} 
As shown in Figure \ref{fig:supp}, the enhancement or suppression modules would collapse to become indistinguishable for different speakers without $L^{Enhance}_{triple}$ and $L^{Suppress}_{triple}$. This emphasizes the necessity of $L^{Enhance}_{triple}$ and $L^{Suppress}_{triple}$ to ensure the enhancement and suppression module effectively capture and differentiate the characteristics of individual speakers.

\noindent\textbf{Visualization Analysis of Feature Suppression.}
In our main submission, we primarily presented the visualization of enhancement weights. However, in this supplementary material, we provide the visualization of suppression weights, which exhibit consistent behavior with the enhancement weights. As shown in left side of Figure \ref{fig:supp}, the visualization of suppression weights demonstrates similar patterns to the enhancement weights.

Specifically, when visualizing the enhancement weights for the same channel, we observe significant differences across different speakers. Similarly, this pattern becomes even more pronounced when examining the suppression weights( Three channels are randomly selected from the set of 64 channels as examples). In some cases, a specific region may undergo significant suppression for one speaker, while the suppression in the same region for another speaker may not be as prominent.

This consistent behavior between the visualization of enhancement and suppression weights further supports the effectiveness of our approach. It indicates that the model effectively learns to enhance content-dependent information and suppress content-independent information in a discriminative manner. 
 
\noindent\textbf{Visualization Analysis of Speaker Features.}
In order to gain a clearer understanding of the speaker features extracted by our model, we conducted a visualization analysis using t-SNE dimensionality reduction. Specifically, we visualized the speaker features for each speaker in the LRW-ID dataset, and we also associated each cluster with the appearance of speakers in the LRW-ID test set, as shown in Figure \ref{fig:tsne}.
\bibliography{egbib}